\documentclass[11pt]{paper}
\usepackage{pslatex}
\usepackage{graphicx}
\usepackage{epsfig}

\begin{document}

\title{{\it AGNOSCO} -\\ Identification of Infected Nodes with artificial Ant Colonies}

\author{
{\bf Michael Hilker and Christoph Schommer}\\
{University of Luxembourg, Campus Kirchberg}\\
{Dept. of Computer Science and Communication}\\
{6, Rue Richard Coudenhove-Kalergi, L-1359 Luxembourg}\\
{Email: \{michael.hilker, christoph.schommer\}@ uni.lu}\\
{Phone: +352-466644-5-\{311,228\}}
}

\maketitle
\thispagestyle{empty}

{\bf Abstract:}
{\it If a computer node is infected by a virus, worm or a backdoor, then this is a security risk for the complete network structure where the node is associated. Existing Network Intrusion Detection Systems (NIDS) provide a certain amount of support for the identification of such infected nodes but suffer from the need of plenty of communication and computational power. In this article, we present a novel approach called {\it AGNOSCO} to support the identification of infected nodes through the usage of artificial ant colonies. It is shown that  {\it AGNOSCO} overcomes the communication and computational power problem while identifying infected nodes properly.}


\section{Motivation}
\label{secIntroduction}
In the current working and life environment, connected nodes - computers, servers, etc. - are essential. These nodes are under constant assault form attacks like e.g. worms, trojans, and hackers. Nowadays, there exist several approaches to protect a computer node or a network against criminal attacks like virus- and malware guards, symbolic NIDS-solutions like SNORT \cite{Roe99,Deb98,Sna91}, and bio-inspired NIDS-solutions (Artificial Immune Systems, \cite{Hof99,Hof00,Spa00}). These protection-systems check each packet, which traverses a network node, and evaluate if this packet intends to attack or not.
However, many NIDS solutions suffer from identifying (new) attacks as well as from the need of plenty of computational power; furthermore, there exist applied techniques to camouflage attacks in a way that NIDS are not able to identify the attack at all. Hence, there are situations when an attack infects a node and when a computer network risks to be infected by the node. This is much more critical as it seems since infections can cause a backdoor to other attacks, infections can send packets containing an attack to infect healthy nodes.

The identification of such an infected node - sometimes also zombie-node called - is a well-know problem. In the current research community, only a few approaches of identifying infected nodes are known, for example
\begin{itemize}
\item	\textit{Anomaly Detection:}  A system knows how to identify normal network traffic and tries to identify abnormal network traffic using this information. A node, which transmits a lot of abnormal traffic, is infected with a high probability.
\item	\textit{Statistically Analysis of Network Traffic:} A system observes the network traffic and if some statistically parameters are met, the node is probably infected. 
\item	\textit{Inference from Network Traffic Analysis:} If a network node is infected, the network node releases several packets containing an attack in order to infect also other nodes of the network. This behaviour can be recognized using intrusion detection and an intelligent inference system is used in order to derive to the infected node.
\item \textit{Trust or the Byzantine General Problem \cite{Lam82}:} If a node runs a service, a watchdog can use this service regularly in order to check for incorrect answers and the watchdog can observe if the node sends packets if it should not in order to detect abnormal behaviour.
\end{itemize}

Unfortunately, all these approaches have significant disadvantages. First, they need information from the computer network that must be collected, fusioned, and further processed. Consequently, this results in high communication costs where the centric evaluation affords plenty of computational power. Second, the last approach shares several other disadvantages, e.g., defining an incorrect answer and deciding when a node should not send any packets. Following this, our motivation is that novel (bio-inspired) systems can significantly contribute to a higher identification rate of infected nodes.

\section{Description of the idea}
\label{secDescribtion}

In this article, we introduce a novel approach called {\it AGNOSCO} that is an acronym for {\it AGents for the ideNtification of infected nOdes uSing artificial ant COlonies}. The advantages of {\it AGNOSCO} are that it works autonomously, distributed and more efficiently. {\it AGNOSCO} does not need either much additional computational power or much additional communicative time while identifying infected nodes properly. 

{\it AGNOSCO} is a part of our implemented network intrusion detection system called {\it SANA} (= Security anAlysis in iNternet trAffic), which is an artificial immune system that uses lightweight, autonomous and adaptive artificial cells for the protection of networks. {\it SANA} is a library of non-standard approaches for network security and compounds these approaches in an artificial immune system. 

To understand {\it AGNOSCO}, we shortly have to introduce the concept of ant colonies. Generally, ants have two states and if an ant carries out a prey, the ant will release a lot of pheromones while traveling back to the ant-hill. This is, because other ants should find this prey as well. However, if an ant does not carry out a prey, it releases only a few or no pheromones. If now an ant navigates, it uses the pheromones determined on the ground. In this respect, ant colonies are used for a lot of computer science problems, e.g. optimization \cite{Gro90,Col91,Dor96}.

For {\it AGNOSCO}, each connection of the network contains a pheromone-value that is increased if an packet with attack travels over the connection; it is decreased, if a packet without attack travels over the connection. The pheromone value is calculated by the application of an affinity-function. This pheromone-value represents the rate of attacked packets over the connection. {\it AGNOSCO} flows through the network, interprets the pheromone value and identifies the infected nodes. Thereafter, a disinfecting-process can be started for this infected node; this disinfection-process is not part of {\it AGNOSCO} which just identifies the infected nodes.

\section{Implementation Details}
\label{secImplementationDetails}
 So far, {\it AGNOSCO} is implemented in {\it SANA} as one of the lightweighted, adaptive and autonomous artificial Cells that flow through the network to protect the computer network. In {\it SANA}, there exist artificial Cells and other components that evaluate packets whether they contain an attack or not \cite{Hil06}. Furthermore, if a packet arrives at the destination, the node confirms this event using a small confirmation-packet which is sent from the destination to the source. If an artificial Cell or another component identifies a packet as malicious, the node will send a confirmation-packet as well. However, this confirmation-packet does not inform the source, it increases the pheromone-level on each connection on its way back to the source-node. The confirmation-packet for a packet without an attack behaves like an ant without a prey and the confirmation-packet for a packet with an attack behaves like an ant carrying out a prey. 
 
For setting the right value of pheromones, the following affinity-function $af$ is chosen:  

$$af_{connection}=\sum_{i=1}^{b}{inc*dec^{\mbox{\#}good\mbox{-}packets_i}} \mbox{\hspace{0.5cm} $($$foreach$ $connection$ $in$ $the$ $network$$)$}$$

where $b$ is the number of infected (bad) packets over this connection and the parameter $inc$ the increasing-factor of the system. The parameter $dec$ is the decreasing-factor of the system and \#$good\mbox{-}packets_i$ the number of good packets which travelled over the connection after the $i$-th bad packet. In this test simulation, we adjusted $inc$ to the value of 20 and $dec$ permanently to 0.95. 

Then, the workflow of the affinity-function is as follows:
\begin{itemize}
\item	If a packet contains no attack travels over a connection, all \#$good\mbox{-}packets_i$ are increased by $1$ because \#$good\mbox{-}packets_i$ is the counter how many good packets travelled over the connection after the $i$-th bad packet.
\item	If a packet contains an attack travels over a connection, a new summand is added to the sum. Hence, the parameter $b$ is increased by $1$ and \#$good\mbox{-}packets_b$ - the counter of good packets of the new bad packet - is set to $0$.
\end{itemize}
Consequently, this affinity-function for a connection increases heavily if a bad-packet is found and decreases slightly if a good packet travels over the connection. We refer for an analysis of this function to the section \ref{secResultAnalysis}.

If a node is infected, it will normally send a high number of packets containing the attack in order to infect other nodes in the computer network. Therefore, there will originate pheromone-tracks in the system which point towards the infected nodes; as already mentioned before, {\it AGNOSCO} identifies the infected nodes. 

{\it AGNOSCO} evaluates the pheromone-level of connections as an artificial cells of the artificial immune system  {\it SANA} while flowing through the computer network: if a pheromone-level is higher than the threshold of  {\it AGNOSCO}, it follows the track. However, if {\it AGNOSCO} follows a track and no connection in a node has a pheromone-value higher than the threshold of  {\it AGNOSCO},  {\it AGNOSCO} stops and ends the track. So, {\it AGNOSCO} knows that this node is infected - if and only if the parameters of  {\it AGNOSCO} are set properly; it then informs other components of {\it SANA} for disinfection or isolation of this node. {\it AGNOSCO} behaves like an ant which leaves the ant hill and tries to find a prey using the pheromones. Consequently, the system simulates an artificial ant colony where the preys are infected nodes.

Summary of the Changes in the Network:\\
In the Network Infrastructure must be for each connection some storage-space added in order to store the pheromone-value, at most $10kB$ per connection. The Network Protocols must not be changed and {\it AGNOSCO} is compatible with existing protocols. Essentially, the NIDS-Behaviour concerning identified malicious packets must be changed; if the NIDS identifies a packet as malicious, additionally it must send a confirmation-packet for this bad-packet in order to update the pheromone-values on the path from source to destination. 

\section{Simulation Results}
\label{secResultSim}
Using the implementation of  {\it AGNOSCO}/{\it SANA} for the identification of infected nodes, we tested several scenarios, where different types of attacks have infected network nodes. In all scenarios, {\it AGNOSCO} tried to identify the infected nodes in order to start the disinfection. Generally, {\it AGNOSCO} identified all infected nodes efficiently and produced a good performance for different attack types. Additionally, we determined an appropriate parameter setting. For example, a scenario with $75$ nodes and $3$ artificial Cells of type, {\it AGNOSCO} took about $30$-$50$ processing-steps to identify all infected nodes. If a new infection of a node occured, {\it AGNOSCO} identified this infected node using at most $20$ processing time-steps.

In the simulation, we found out a certain amount of computational power and the communication bandwidth that is requested by {\it AGNOSCO}. The computational power is nearly not recognisable since {\it AGNOSCO} flows through the network and reads and rates just the pheromones values. For the storing of the pheromone-value in each connection, only storage-space was needed that is limited to $10$kB. The communication bandwidth depended on how many artificial Cells of this type are flowing through the network. A simulation of $75$ nodes and $3$ artificial Cells of type {\it AGNOSCO} requested only little storage and one or two IP-Packets per time-step. Consequently, hardware requirements can be met by standard computer networks while additional communication does not matter in common used high-speed networks. 

\section{A closer look to the Affinity-Function}
\label{secResultAnalysis}

\begin{figure}[t]
  \begin{minipage}{0.48\textwidth}
    \begin{center}
      \fbox{\epsfig{file=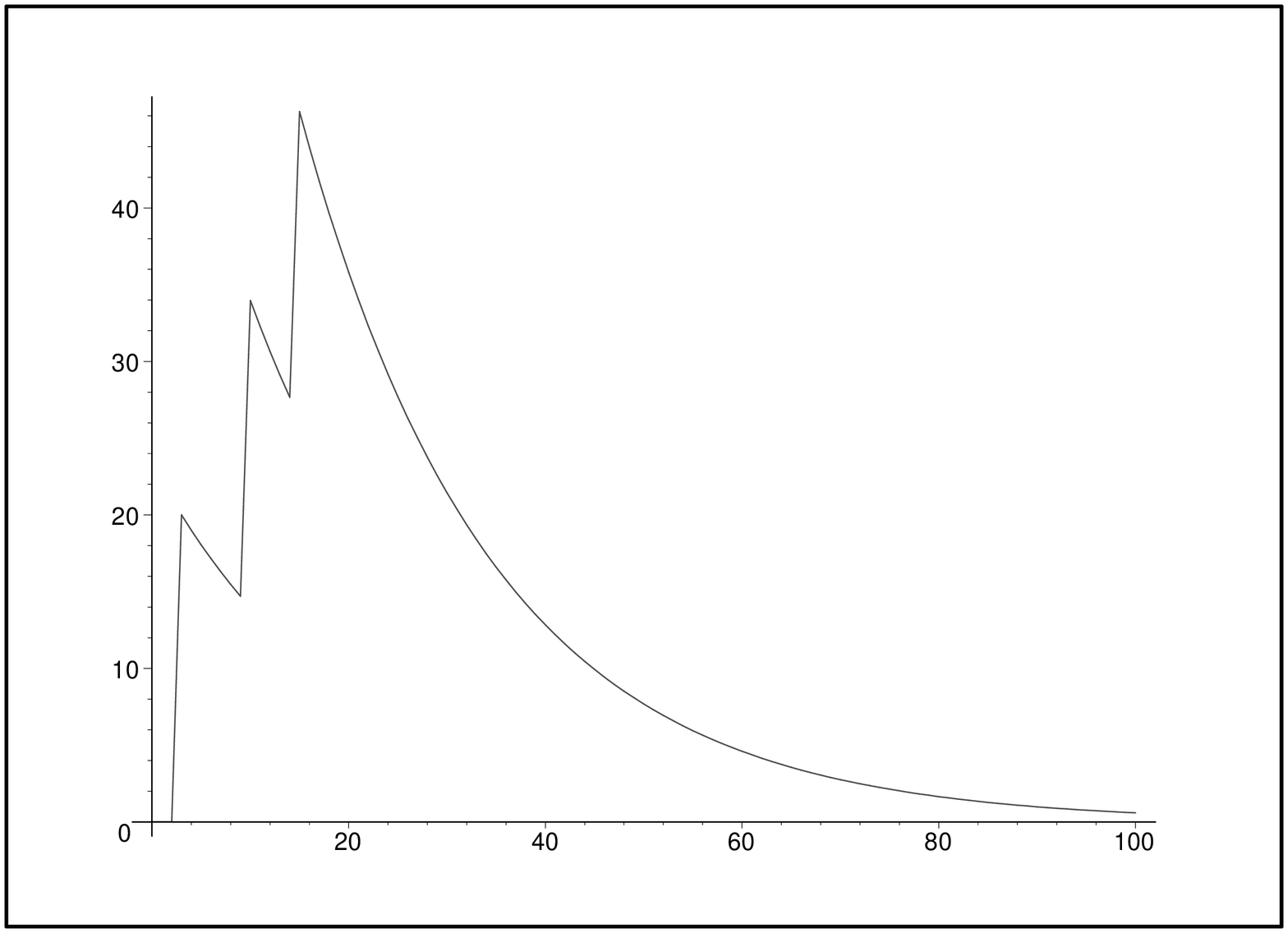, height=6.5cm, angle=270}}
      \caption{A short-term plot of the affinity-function used in our simulations.}
      \label{AffinityFunctionPlot}
    \end{center}
  \end{minipage}
  \begin{minipage}{0.04\textwidth}
     \hfill 
  \end{minipage}%
  \begin{minipage}{0.48\textwidth}
    \begin{center}
      \fbox{\epsfig{file=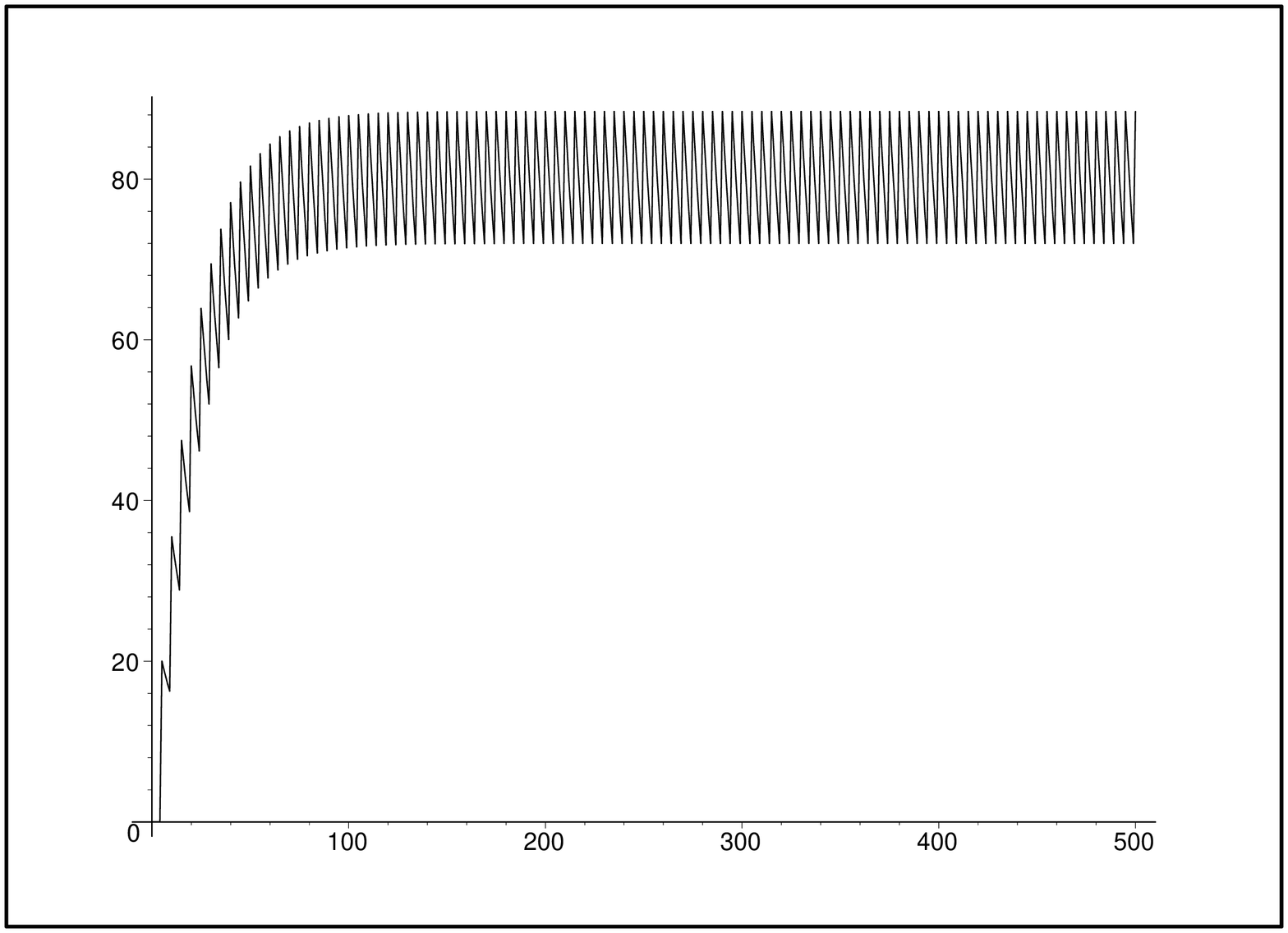, height=6.5cm, angle=270}}
      \caption{A long-term plot of the affinity-function used in our simulations.}
      \label{AffinityFunctionLongPlot}
    \end{center}
  \end{minipage}
\end{figure}


The affinity-function is biologically inspired. In human affinity-functions, an event increases the affinity heavily and, over time if no new event occurs, the value of the affinity-function decreases primarily heavily and afterwards slowly. This means, that the gradient of the function is primarily high and decreases afterwards. Thus, the human body reacts using the affinity-function to an event heavily; thereafter, with the high gradient, the human body tries to compensate an error; and afterwards, with the low gradient, it tries to reach a stable value. 

Consequently, the affinity-function tries to model the behaviour that the value increases by leaps and bounds in order to mark if in a short time-step a lot of bad packets are found for a connection. On the other hand, the affinity-function tries primarily to decrease the value fast in order to eliminate a possible error. Afterwards, the affinity-function value decrease slowly in order to reach a stable value so that pheromone-tracks appear in the network. The affinity-function forgets also old events of bad-packets in order to model the behaviour of the human affinity-function and in order to reach a stable value. 

Figure \ref{AffinityFunctionPlot} visualizes a short-term plot of the affinity-function. The parameters are $inc=20$, $dec=0.95$ and the number of packets is $100$ and the packets number $3$, $10$ and $15$ are identified as bad and all other packets are good. 

Figure \ref{AffinityFunctionLongPlot} visualises the long-term behaviour of the affinity-function if an infected node is nearby the connection. Every fifth packet is an identified bad-packet and all other packets are good. The parameters $inc=20$ and $dec=0.95$ are equal to the last plot of the affinity-function. The figure shows that the value of the affinity-function straight fast towards a stable value and the value alternates around it. 

With this affinity-function, the pheromone-tracks appear in the network properly and the artificial Cells can follow these tracks as well as identify the infected nodes; in our simulations the threshold of {\it AGNOSCO} was $10$. 

\section{Conclusion}
\label{secConclusion}
In this article, we described a novel approach called {\it AGNOSCO} for the identification of infected nodes in a computer network. The idea is biologically inspired and follows the behaviour of ant colonies. {\it AGNOSCO} is implemented, simulated and tested; {\it AGNOSCO} efficiently identifies the infected network nodes unless taking both additional computational power and additional communication bandwidth. We are sure that {\it AGNOSCO} can enhance commonly used NIDS as well as {\it SANA}. Future enhancements of {\it SANA} especially the communication and collaboration of the artificial Cells in SANA will be our next challenges. 

\section*{Acknowledgments}
 {\it SANA} and {\it AGNOSCO} are part of the project INTRA (= INternet TRAffic management and analysis) that are financially supported by the University of Luxembourg. We would like to thank the Ministre Luxembourgeois de l'education et de la recherche for additional financial support and Jacob Zimmermann (Queensland University of Technology) for worthful discussions.

\bibliography{Paper}
\bibliographystyle{plain}

\end{document}